\documentclass[letterpaper]{article} 
\usepackage[preprint]{preprint}
\usepackage[hyphens]{url}  
\usepackage{graphicx} 
\usepackage{natbib}  
\usepackage{caption} 
\usepackage{algorithm}
\usepackage{algorithmic}
\usepackage{booktabs}
\usepackage{amsmath}
\usepackage{amssymb}
\usepackage{multirow}
\usepackage{colortbl}
\usepackage{pifont}

\newcommand{\method}{Codebook Agent}

\definecolor{darkgray}{rgb}{0.5,0.5,0.5}
\definecolor{lightpurple}{rgb}{0.9,0.9,1.0}
\definecolor{ForestGreen}{HTML}{228B22}
\newcommand{\greenup}[1]{\textcolor{ForestGreen}{\footnotesize\raisebox{0.8pt}{$\scriptstyle\blacktriangle$}{#1}}}
\newcommand{\reddown}[1]{\textcolor{red}{\footnotesize\raisebox{0.8pt}{$\scriptstyle\blacktriangledown$}{#1}}}

\title{Codebook Agent: Amortized Topology Design for LLM Multi-Agent Systems}
\author{
    Jinxi Yu\equalcontrib,
    Yubei Li\equalcontrib,
    Eric Hanchen Jiang\equalcontrib,
    Zhi Zhang,
    Dong Liu,\\
    Wenxiao Zhao,
    Levina Li,
    Kai-Wei Chang,
    Ying Nian Wu
}
\affiliations{University of California, Los Angeles}

\begin{document}

\maketitle

\begin{abstract}
Adapting the communication topology of an LLM multi-agent system to each query improves both accuracy and efficiency, yet current designers treat this as conditional graph generation: a variational, autoregressive, or diffusion decoder searches the $N{\times}N$ adjacency space, and a graph-network proxy trained on utility and a structural cost such as edge count ranks the sampled candidates. We argue that this formulation is misaligned with the problem. Empirically, topologies that survive a reward filter collapse to about six distinct graphs even when the codebook capacity grows from 8 to 64; edge count is \emph{negatively} correlated with measured token consumption (Pearson $r \approx -0.4$), so sparsifying the graph makes inference more expensive; and a message-passing scorer over agent-profile nodes is adjacency-invariant whenever agents share a profile---the default configuration of published benchmarks---so it cannot rank candidates at all in that regime. These three facts motivate \emph{Codebook Agent}: a vector-quantized autoencoder compresses successful topologies into a query-independent 16-entry codebook; a reward-weighted MLP maps the query embedding to a distribution over codes; and an MLP proxy that reads the flattened adjacency, regressed on measured utility and per-task normalized token cost, reranks the top decoded candidates in a single batched forward pass. With no iterative search and no message passing at test time, Codebook Agent is the most accurate method on all six benchmarks we compare (84.6 average against 83.0 for the strongest prior designer), emits a topology in 2.4\,ms, and uses 21.9--33.2\% fewer LLM tokens. Our code is available here: \url{https://github.com/jinxiy1104/CodebookAgent}.
\end{abstract}

\begin{figure}[t]
\centering
\includegraphics[width=\columnwidth]{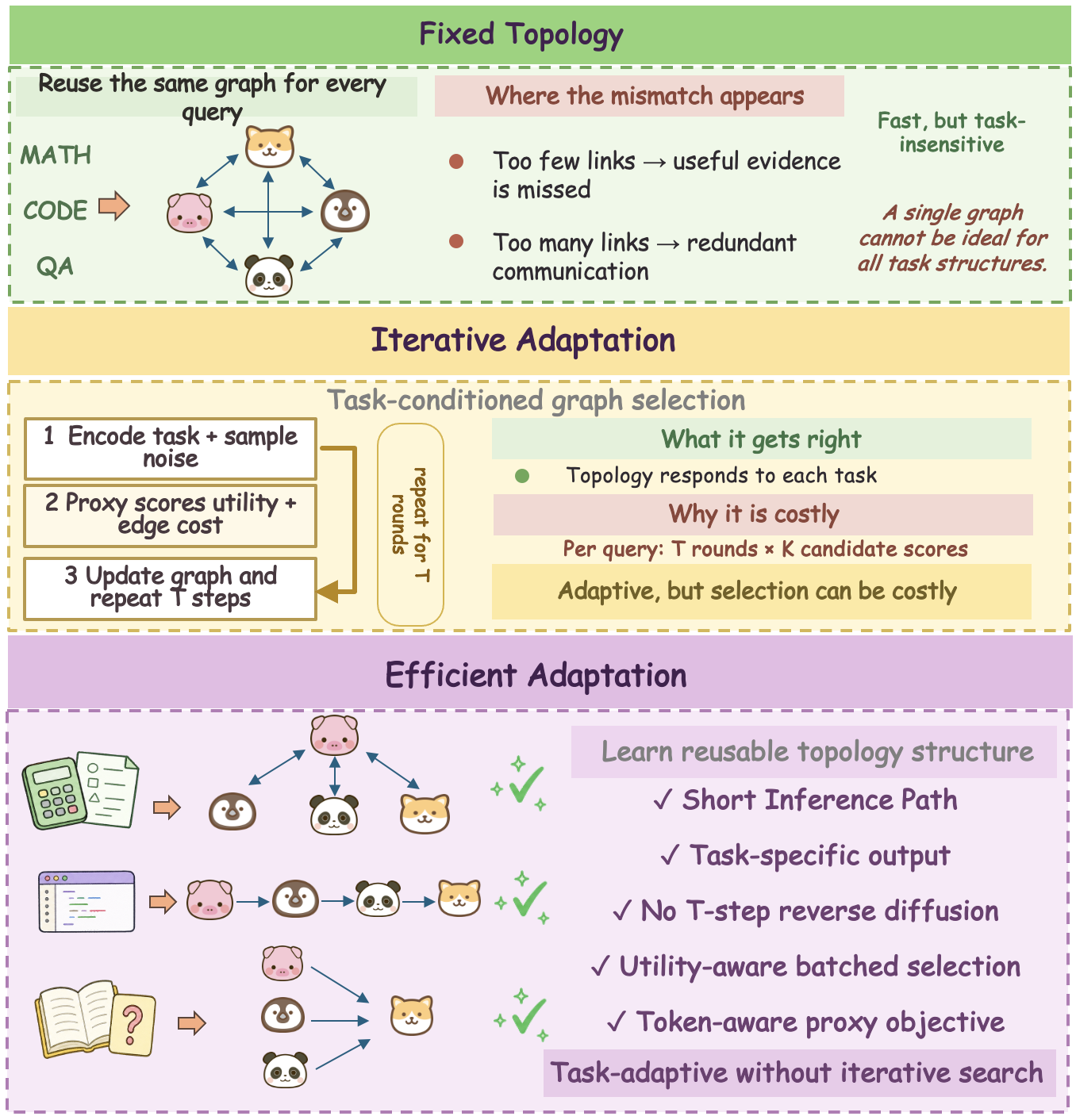}
\caption{\textbf{Three regimes of topology design.}
(Top)~A fixed topology is reused for every query: zero design cost, but no single graph fits every task.
(Middle)~Query-conditioned generators adapt the graph via an iterative loop that samples and scores $K$ candidates at each of $T$ steps, typically with a proxy trained on utility and edge count.
(Bottom)~Codebook Agent keeps per-query adaptation and removes the loop: the design space is discretized once offline into a codebook, a query embedding selects codes, and one batched proxy call under a token-based cost objective returns the winner.}
\label{fig:intro}
\end{figure}

\section{Introduction}

Multi-agent systems built from LLM agents solve reasoning, question answering, and coding tasks by exchanging intermediate outputs over a communication topology \citep{wu2024autogen,hong2024metagpt,qian2024chatdev,du2024debate}. That topology is not bookkeeping: it decides which agents see which messages, and both accuracy and token consumption move with it \citep{qian2025macnet,zhuge2024gptswarm,zhang2025agentprune}. The field has therefore progressed from hand-designed structures to learned ones, and most recently to \emph{query-conditioned} generators that emit a fresh topology for every input \citep{zhang2025gdesigner,jiang2026gtd}.

The latest designers share a natural but costly recipe. A variational \citep{zhang2025gdesigner}, autoregressive \citep{li2026argdesigner}, or diffusion decoder \citep{jiang2026gtd} searches the $N{\times}N$ adjacency space; a graph network over agent-profile nodes then ranks candidates with a utility head and a structural cost head, typically the edge count $|E|$ \citep{zhuge2024gptswarm,zhang2025agentprune,zhang2025gdesigner,jiang2026gtd}. The recipe looks principled---generate, then score---yet it quietly assumes that the useful design space is large enough to need a generative decoder, that $|E|$ tracks inference cost, and that message passing can tell two topologies apart. Our measurements reject all three assumptions.

Figure~\ref{fig:intro} sketches the consequence. First, the useful design space is a short list, not a manifold: topologies that solve their task fall into about six distinct codes even as codebook capacity grows from 8 to 64, and the best fixed topology stays within 1.4 accuracy points of every generated one we measure---so capacity spent modeling adjacency is spent where the problem does not live. Second, the structural cost surrogate is inverted: edge count correlates with measured tokens at $r \approx -0.4$, because sparse communication yields longer completions, so minimizing $|E|$ \emph{maximizes} the bill it was introduced to reduce. Third, on the homogeneous teams that dominate published benchmarks, a message-passing scorer over profile nodes is adjacency-invariant: every candidate receives the same score, the ranking does not exist, and hundreds of guided diffusion steps reproduce a constant.

If the problem is to choose among a few graphs under a measured cost, the matching designer is not another decoder. \emph{Codebook Agent} therefore amortizes topology design into three feed-forward pieces: a vector-quantized autoencoder \citep{oord2017vqvae} that indexes successful topologies into 16 codes, a reward-weighted MLP that maps the query embedding to a distribution over those codes, and an MLP proxy that reads the flattened adjacency and is regressed on measured utility and per-task normalized token cost. At test time we decode the top codes, score them in one batched proxy call, and execute the winner---2.4\,ms, with no sampling loop and no message passing.\\ \textbf{Our contributions}:
\begin{itemize}
\item[\ding{182}] We characterize topology design rather than assume its form: the reward-surviving space collapses to about six graphs independently of model capacity, the common structural cost surrogate is inverted w.r.t.\ measured tokens, and message-passing proxies are topology-blind on homogeneous teams.
\item[\ding{183}]  We propose Codebook Agent, the minimal designer these facts admit, a discrete codebook, a reward-weighted code predictor, and one batched pass of an execution-grounded proxy, with no iterative search and no message passing.
\item[\ding{184}]  On six benchmarks and two LLM backends, against single-agent prompting, multi-agent collaboration, and learned topology designers, Codebook Agent is the most accurate method on every benchmark of Table~\ref{tab:main}, generates topologies in 2.4\,ms, and reduces token consumption by 21.9--33.2\% (Tables~\ref{tab:main} and~\ref{tab:new-results}).
\end{itemize}

\section{Related Work}

\paragraph{Communication topologies for LLM agents.}
\looseness=-1
Early frameworks fix the topology by hand \citep{hong2024metagpt,qian2024chatdev,wu2024autogen,chen2024agentverse,du2024debate}. A second line optimizes it: GPTSwarm learns edge distributions with policy gradients \citep{zhuge2024gptswarm}, DyLAN selects agents dynamically \citep{liu2024dylan}, search-based systems explore agentic designs offline \citep{hu2025adas,zhang2025aflow}, pruning methods sparsify a fixed structure to save tokens \citep{zhang2025agentprune,wang2025agentdropout}, and Optima trains the agents for efficiency \citep{chen2025optima}. Closest to us are query-conditioned generators \citep{zhang2025gdesigner,jiang2026gtd}. We share their problem statement, one topology per query, and differ in what we assume about it: they model the adjacency space as something to be generated and scored by a graph network, and we show it is a short list to be selected from and scored by measured cost.

\paragraph{Graph generation and discrete latents.}
Deep graph generators are predominantly iterative, whether autoregressive \citep{you2018graphrnn} or score-based \citep{jo2022gdss,vignac2023digress,zhou2024lgd}, inheriting a sampling cost \citep{ho2020ddpm} that faster samplers reduce but do not remove \citep{song2021ddim}; one-shot VAE decoders exist for small graphs \citep{simonovsky2018graphvae}. We need one small graph per query under a latency constraint, so we discretize the output space with vector quantization \citep{oord2017vqvae,razavi2019vqvae2} instead of sampling from a continuous process. VQGraph tokenizes local structure to improve GNN-to-MLP distillation \citep{yang2024vqgraph,zhang2022glnn,tian2023nosmog,wu2023ffg2m,hinton2015distillation}; we borrow its structure-token regularizer, but our proxy has no teacher to distill from: on homogeneous teams a message-passing teacher is constant per query, so we train the MLP directly on measured rewards.

\paragraph{Cost of multi-agent inference.}
Multi-agent pipelines multiply LLM calls, and returns diminish or invert as calls grow \citep{li2024moreagents,chen2024scaling,smit2024mad}, with many observed failures attributable to coordination rather than single-agent ability \citep{cemri2025mast}. Cost-aware model selection addresses the single-call case \citep{chen2024frugalgpt}. These results motivate treating measured tokens, not a structural surrogate, as the cost objective of topology design.

\begin{figure*}[t]
\centering
\includegraphics[width=\textwidth]{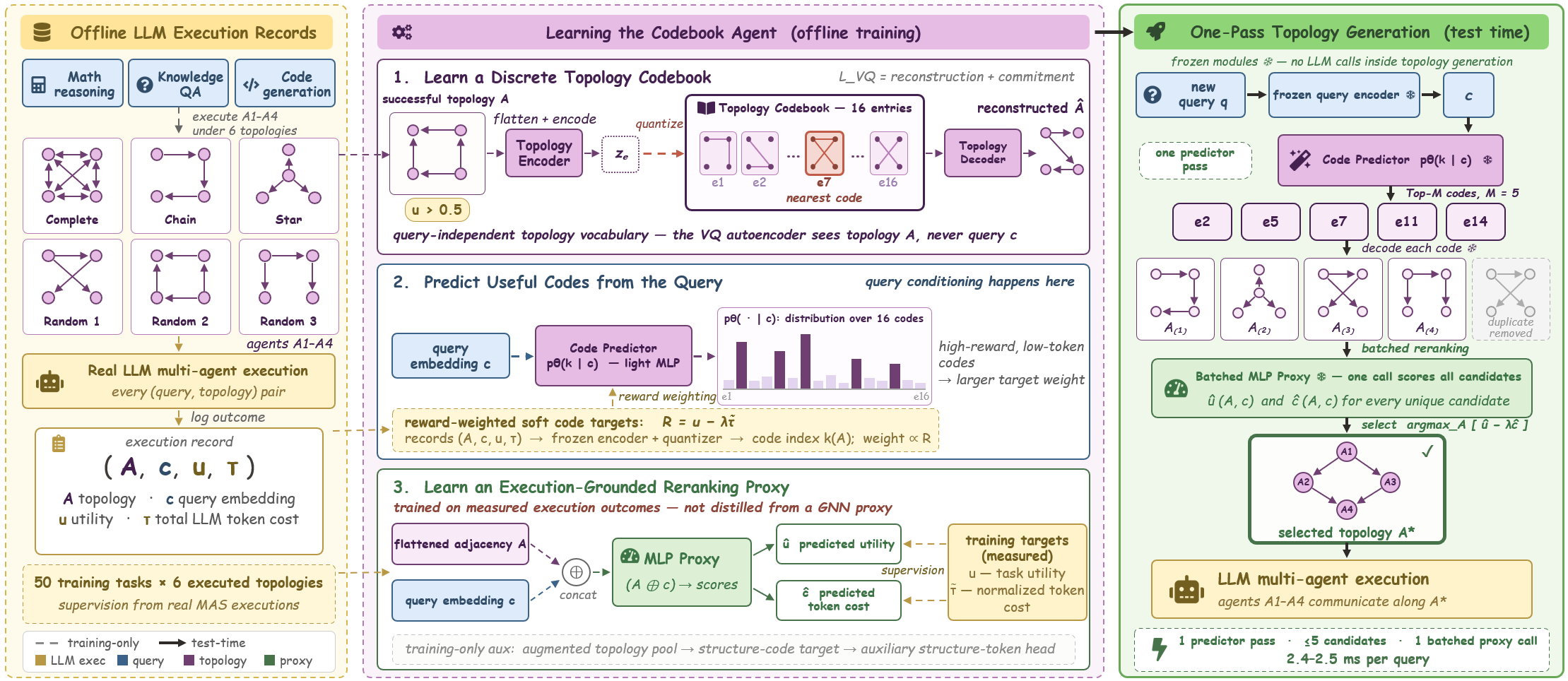}
\caption{\textbf{Overview of Codebook Agent.}
\emph{Left:} offline collection executes fixed topologies on the training split and logs $(A,c,u,\tau)$ for every (query, topology) pair.
\emph{Middle:} (1)~a VQ autoencoder compresses successful topologies ($u>0.5$) into a query-independent codebook; (2)~an MLP predictor $p_\theta(k\mid c)$ is fit to the reward-weighted soft targets of Eq.~\eqref{eq:soft}; (3)~an MLP proxy $f_\phi$ is regressed on measured utility and per-task normalized token cost (Eqs.~\eqref{eq:proxy}--\eqref{eq:proxy-code}), with an auxiliary structure-token head used in training only.
\emph{Right:} at test time the top $M$ codes are decoded, deduplicated, and scored in one batched proxy call; the candidate maximizing $\hat{u}-\lambda\hat{c}$ (Eq.~\eqref{eq:infer}) is executed.
Dashed arrows are training-only; solid arrows are the frozen test-time path (no LLM calls inside topology generation).}
\label{fig:overview}
\end{figure*}

\section{Problem Setup and Background}
\label{sec:problem}

\paragraph{Topology design.}
A team of $N$ agents has profile texts $p_1, \dots, p_N$ embedded as $x_i = E(p_i) \in \mathbb{R}^{d}$ by a frozen sentence encoder ($d = 384$). A topology is a directed adjacency matrix $A \in \{0,1\}^{N \times N}$ without self-loops, where $A_{ij} = 1$ shows the output of agent $i$ to agent $j$, and a decision node aggregates the final answers following GDesigner \citep{zhang2025gdesigner}. Executing the team on query $q$ under $A$ returns a utility $u(A, q) \in \{0, 1\}$ and a token cost $\tau(A, q)$. Writing $c = E(q)$ for the query embedding, the goal is a generator mapping $c$ to a topology maximizing $R = u - \lambda \tilde{\tau}$, with $\tilde{\tau}$ a normalized cost and $\lambda = 0.1$. All methods in this paper are trained on the same records: per benchmark, 50 training tasks executed under 6 fixed topologies (complete, chain, star, and three Erdos-Renyi samples) with real LLM agents, giving 300 tuples $(A_j, c_j, u_j, \tau_j)$.

\paragraph{Two design axes.}
A query-conditioned designer is determined by two independent choices, which we vary separately in the experiments. \emph{Axis A, the candidate generator}, maps $c$ to one or more topologies; the incumbent choice is an iterative or continuous decoder over the adjacency space, variational \citep{zhang2025gdesigner}, autoregressive \citep{li2026argdesigner}, or a diffusion reverse process with $T$ steps and $K$ candidates per step \citep{jiang2026gtd}, which we measure at 301 to 396 ms per query. \emph{Axis B, the scorer}, ranks candidates; the incumbent choice, shared across the learned-design literature \citep{zhuge2024gptswarm,zhang2025agentprune,zhang2025gdesigner,jiang2026gtd}, is a message-passing network over a graph whose nodes carry the profile embeddings $x_i$ and whose edges are the candidate $A$:
\begin{equation}
\label{eq:incumbent}
f_{\mathrm{gnn}}(A, c) = h\!\left( \mathrm{pool}\big( \mathrm{MPNN}(A, \{x_i\}, c) \big) \right) \in \mathbb{R}^2,
\end{equation}
with a two-dimensional head regressed onto utility and a structural cost label, most commonly the edge count $|E|$. We implement Eq.~\eqref{eq:incumbent} as a GAT \citep{velickovic2018gat} and treat it as one level of Axis B, so it can be swapped into our own pipeline with everything else held fixed. The introduction's three observations---a collapsed design space, an inverted cost surrogate, and adjacency-blind message passing on homogeneous teams---determine how we instantiate both axes below.

\section{Method}

Codebook Agent instantiates Axis~A as a discrete codebook with a query-conditioned prior, and Axis~B as an MLP over flattened adjacencies trained on measured utility and token cost. Let $\mathcal{D}=\{(A_j,c_j,u_j,\tau_j)\}_{j=1}^{|\mathcal{D}|}$ be the execution records of Section~\ref{sec:problem}. Define the composite reward
\begin{equation}
\label{eq:reward}
R_j \;=\; u_j - \lambda\,\tilde{\tau}_j,
\qquad
\tilde{\tau}_j \;=\; \frac{\tau_j}{\displaystyle\tfrac{1}{|\{j':c_{j'}=c_j\}|}\sum_{j':c_{j'}=c_j}\tau_{j'}},
\end{equation}
with $\lambda=0.1$: $\tilde{\tau}_j$ is the token count normalized by the mean token count of the same task, so difficulty cancels and within-task rankings remain. Using measured tokens rather than $|E|$ is essential: on our records edge count correlates with $\tau$ at $r\approx -0.4$, so an edge-count head pushes the system toward more expensive graphs. The codebook, code predictor, and reranking proxy are trained on $\mathcal{D}$ with no further LLM calls. Figure~\ref{fig:overview} shows offline collection, the three training stages, and one-pass test-time generation.

\subsection{Topology Codebook}

If reward-surviving topologies occupy only a handful of distinct graphs, the generator should index that short list rather than search the adjacency manifold. We therefore compress observed topologies with a vector-quantized autoencoder. Flatten the off-diagonal entries $a=\mathrm{vec}(A)\in\{0,1\}^{N(N-1)}$ and encode
\begin{equation}
\label{eq:encode}
\begin{aligned}
z_e &= \mathrm{Enc}_\psi(a) \in \mathbb{R}^{d_z},
\qquad
k^\star = \arg\min_{k\in\{1,\dots,K\}} \bigl\| z_e - e_k \bigr\|_2, \\
z_q &= e_{k^\star}.
\end{aligned}
\end{equation}

with $d_z=32$ and codebook size $K=16$. The codebook $\{e_k\}_{k=1}^{K}$ is updated by exponential moving averages (decay $0.99$), with commitment coefficient $\beta=0.25$, straight-through gradients, and reinitialization of codes unused for 50 batches \citep{oord2017vqvae,razavi2019vqvae2}. A mirror-image decoder maps $z_q$ to edge logits $\hat{\ell}=\mathrm{Dec}_\omega(z_q)\in\mathbb{R}^{N(N-1)}$; the diagonal is fixed to zero and the graph remains directed. Writing $\hat{a}=\sigma(\hat{\ell})$, the training objective is
\begin{equation}
\label{eq:vq}
\mathcal{L}_{\mathrm{vq}}
\;=\;
\mathrm{BCE}_{w}\!\bigl(\hat{a},\,a\bigr)
\;+\;
\beta\,\bigl\| z_e - \mathrm{sg}[z_q] \bigr\|_2^2,
\end{equation}
where $\mathrm{BCE}_{w}$ weights positive entries by the zero-to-one ratio of the training set to counter edge sparsity and $\mathrm{sg}$ is the stop-gradient. We fit Eq.~\eqref{eq:vq} only on records with $u_j>0.5$, so the codebook stores topologies that solved their task. Neither encoder nor decoder is conditioned on the query: after training, code $k$ decodes to the binary topology
\begin{equation}
\label{eq:decode}
A^{(k)} \;=\; \mathbf{1}\bigl[\sigma\bigl(\mathrm{Dec}_\omega(e_k)\bigr) \ge 1/2\bigr],
\end{equation}
and all query dependence is carried by the predictor below. We train for 200 epochs with Adam at learning rate $3{\times}10^{-4}$ and batch size 16.

\subsection{Reward-Weighted Code Prediction}

At test time there is no target topology to encode, so we learn a conditional prior over codes. Passing every training record through the frozen encoder and quantizer yields indices $k_j$. For each distinct condition $c$ we form the soft target
\begin{equation}
\label{eq:soft}
y_k(c)
\;=\;
\frac{
\sum_{j:\,c_j=c,\,k_j=k}\exp(\gamma R_j)
}{
\sum_{k'=1}^{K}\sum_{j:\,c_j=c,\,k_j=k'}\exp(\gamma R_j)
},
\qquad \gamma=2,
\end{equation}
which concentrates mass on codes whose topologies earned high reward under Eq.~\eqref{eq:reward}, so the prior already prefers cheap, successful graphs. The predictor $p_\theta(k\mid c)$ is an MLP ($d{\to}256{\to}256{\to}K$, dropout $0.1$) trained by soft cross-entropy
\begin{equation}
\label{eq:pred}
\mathcal{L}_{\mathrm{pred}}
\;=\;
-\sum_{c}\sum_{k=1}^{K} y_k(c)\,\log p_\theta(k\mid c).
\end{equation}
This is the only place the query enters generation, so the map from query to candidate topologies is a single matrix pipeline.

\subsection{Execution-Grounded MLP Proxy}

The scorer must (i)~read the adjacency itself and (ii)~regress measured tokens, not $|E|$. The first requirement is forced by the homogeneous-team regime: when $x_1=\dots=x_N=x$, any message-passing layer of Eq.~\eqref{eq:incumbent} aggregates identical features over self-loops, so every node state---and thus the pooled score---is independent of $A$. The same holds for mean and sum aggregators; there is nothing to distill from such a teacher on anonymous teams. We therefore score candidates with an MLP over the flattened adjacency,
\begin{equation}
\label{eq:proxy}
f_\phi(A,c)
\;=\;
\mathrm{MLP}_\phi\bigl(\bigl[\mathrm{vec}(A);\,c\bigr]\bigr)
\;=\;
\bigl(\hat{u},\,\hat{c}\bigr)\in\mathbb{R}^2,
\end{equation}
and regress directly on the measured targets,
\begin{equation}
\label{eq:proxy-reward}
\mathcal{L}_{\mathrm{rew}}
\;=\;
\frac{1}{|\mathcal{D}|}\sum_{j=1}^{|\mathcal{D}|}
\Bigl(
\bigl(\hat{u}_j - u_j\bigr)^2
+
\bigl(\hat{c}_j - \tilde{\tau}_j\bigr)^2
\Bigr).
\end{equation}

Because $|\mathcal{D}|$ covers the topology space thinly, we regularize with a structure-token objective in the spirit of VQGraph \citep{yang2024vqgraph}. A separate reconstruction-only quantizer with $K_{\mathrm{tok}}=32$ codes is fit over an augmented pool $\mathcal{A}^+$ of topologies (static families, Erd\H{o}s--R\'enyi graphs at eight densities, single-edge perturbations, and Bernoulli samples of smoothed edge profiles, capped at 100 per condition). Writing $q(k\mid A)=\mathrm{softmax}_k(-\|z_e(A)-e_k\|_2^2)$ for the soft code assignment of a pool graph, an auxiliary head predicts $q$ under
\begin{equation}
\label{eq:proxy-code}
\mathcal{L}_{\mathrm{code}}
\;=\;
-\mathbb{E}_{A\sim\mathcal{A}^+}\sum_{k=1}^{K_{\mathrm{tok}}}
q(k\mid A)\,\log \hat{q}_\phi(k\mid A),
\end{equation}
and the total proxy loss is $\mathcal{L}_\phi=\mathcal{L}_{\mathrm{rew}}+\lambda_{\mathrm{code}}\mathcal{L}_{\mathrm{code}}$ with $\lambda_{\mathrm{code}}=0.1$. The augmented pool contributes only to Eq.~\eqref{eq:proxy-code}; reward supervision comes exclusively from real executions. The proxy is trained for 200 epochs with Adam at learning rate $10^{-3}$ and batch size 256.

\subsection{One-Pass Inference}

Given a query embedding $c$, let $\mathcal{C}_M(c)=\{k_1,\dots,k_M\}$ be the top $M=5$ codes under $p_\theta(\cdot\mid c)$, and let $\mathcal{A}(c)=\{A^{(k)}:k\in\mathcal{C}_M(c)\}$ after removing duplicates under Eq.~\eqref{eq:decode}. The selected topology is
\begin{equation}
\label{eq:infer}
A^\star(c)
\;=\;
\arg\max_{A\in\mathcal{A}(c)}
\Bigl(\hat{u}(A,c)-\lambda\,\hat{c}(A,c)\Bigr),
\end{equation}
where $(\hat{u},\hat{c})=f_\phi(A,c)$ is evaluated for all survivors in one batched forward pass. Generation therefore costs one predictor pass, at most $M$ decoder passes, and one dense proxy pass---a fixed number of matrix products, with no sampling loop, no sparse graph construction, and no message passing on the test-time path.

\begin{figure}[t]
\centering
\includegraphics[width=0.45\textwidth]{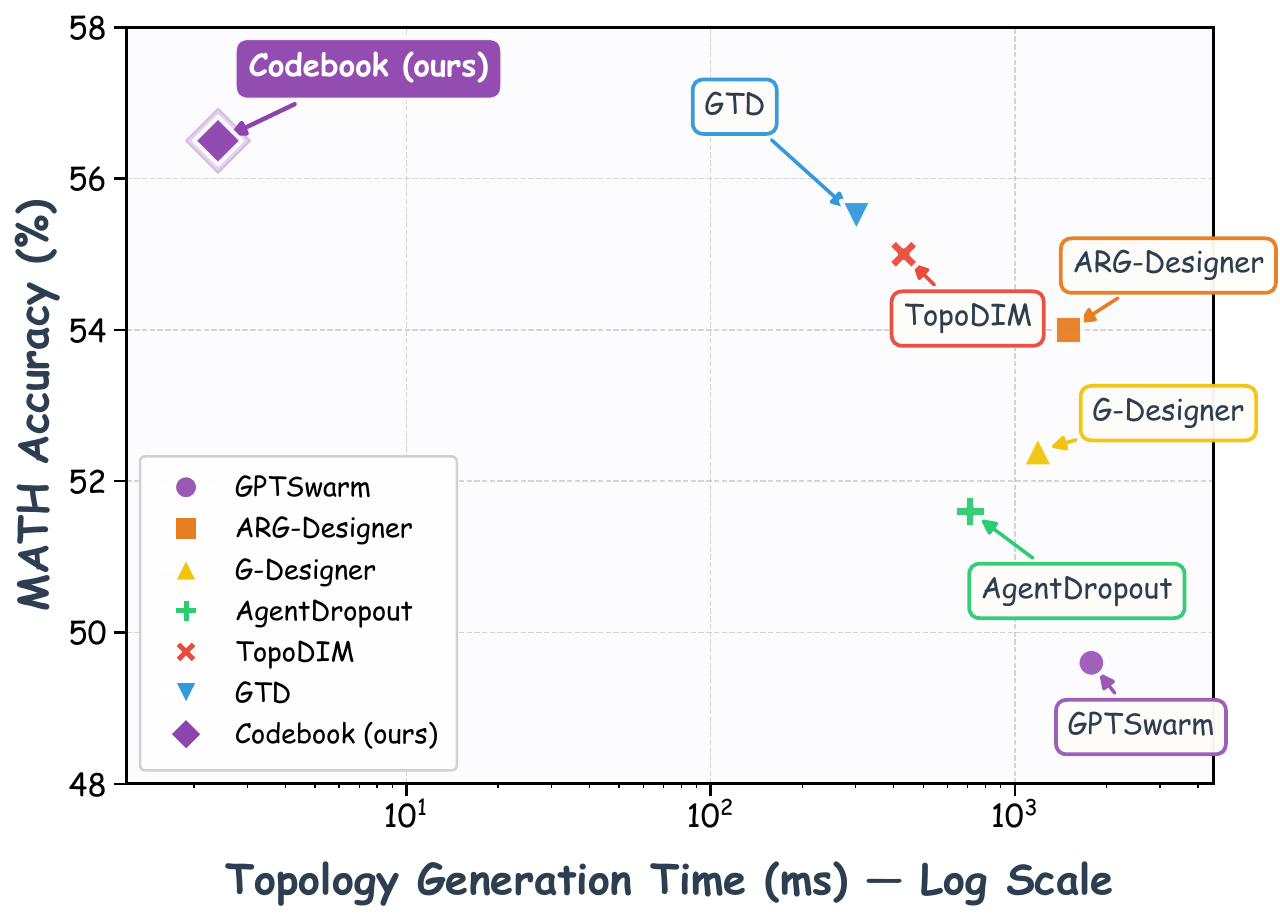}
\caption{\textbf{Accuracy--latency Pareto on MATH.}
Accuracy from Table~\ref{tab:main} versus median topology-generation latency (log-scale $x$-axis). Codebook Agent occupies the upper-left frontier: highest accuracy at 2.4\,ms, against 301--396\,ms for iterative generators.}
\label{fig:efficiency}
\end{figure}

\begin{table*}[t]
\centering
\small

\setlength{\tabcolsep}{4.6pt}
\renewcommand{\arraystretch}{1.10}

\resizebox{\textwidth}{!}{%
\begin{tabular}{lccccccc}
\toprule

\rowcolor{gray!10}
\multirow{2}{*}{\textbf{\textsc{Method}}}
& \multicolumn{4}{c}{\textbf{General/ Math reasoning}}
& \multicolumn{2}{c}{\textbf{Code generation}}
& \multirow{2}{*}{\textbf{Avg.}} \\

\cmidrule(lr){2-5}
\cmidrule(lr){6-7}

\rowcolor{gray!10}
& \textbf{GSM8K}
& \textbf{MATH}
& \textbf{MultiArith}
& \textbf{SVAMP}
& \textbf{MBPP}
& \textbf{HumanEval}
& \\

\midrule

\rowcolor{gray!10}
\multicolumn{8}{l}{\emph{single-agent prompting}} \\

Vanilla
& 87.0
& 47.6
& 97.2
& 88.5
& 72.4
& 73.1
& 77.63 \\

\rowcolor{gray!10}
CoT
& 86.5\reddown{0.5}
& 48.0\greenup{0.4}
& 96.7\reddown{0.5}
& 89.0\greenup{0.5}
& 74.0\greenup{1.6}
& 74.4\greenup{1.3}
& 78.10\greenup{0.47} \\

SC-CoT
& 87.0 \greenup{0.0}
& 49.6\greenup{2.0}
& 97.2 \greenup{0.0}
& 89.5\greenup{1.0}
& 75.0\greenup{2.6}
& 75.0\greenup{1.9}
& 78.88\greenup{1.25} \\

\midrule
\rowcolor{gray!10}
\multicolumn{8}{l}{\emph{multi-agent collaboration}} \\

LLM-Debate
& 89.0\greenup{2.0}
& 50.0\greenup{2.4}
& 97.8\greenup{0.6}
& 90.5\greenup{2.0}
& 76.4\greenup{4.0}
& 75.0\greenup{1.9}
& 79.78\greenup{2.15} \\

\rowcolor{gray!10}
LLM-Blender
& 87.5\greenup{0.5}
& 48.4\greenup{0.8}
& 97.8\greenup{0.6}
& 89.0\greenup{0.5}
& 75.6\greenup{3.2}
& 75.0\greenup{1.9}
& 78.88\greenup{1.25} \\

DyLAN
& 89.5\greenup{2.5}
& 50.0\greenup{2.4}
& 97.8\greenup{0.6}
& 90.5\greenup{2.0}
& 77.0\greenup{4.6}
& 76.3\greenup{3.2}
& 80.18\greenup{2.55} \\

\midrule
\rowcolor{gray!10}
\multicolumn{8}{l}{\emph{learned agent and topology design}} \\

GPTSwarm
& 88.5\greenup{1.5}
& 49.6\greenup{2.0}
& 97.2 \greenup{0.0}
& 90.5\greenup{2.0}
& 76.0\greenup{3.6}
& 75.6\greenup{2.5}
& 79.57\greenup{1.94} \\

\rowcolor{gray!10}
ADAS
& 85.5\reddown{1.5}
& 44.4\reddown{3.2}
& 96.7\reddown{0.5}
& 88.0\reddown{0.5}
& 71.0\reddown{1.4}
& 70.6\reddown{2.5}
& 76.03\reddown{1.60} \\

AFLOW
& 90.5\greenup{3.5}
& 52.4\greenup{4.8}
& 96.7\reddown{0.5}
& 90.0\greenup{1.5}
& 78.4\greenup{6.0}
& 76.9\greenup{3.8}
& 80.82\greenup{3.19} \\

\rowcolor{gray!10}
MaAS
& 91.5\greenup{4.5}
& 53.6\greenup{6.0}
& 97.2\greenup{0.0}
& 91.5\greenup{3.0}
& 79.0\greenup{6.6}
& 76.9\greenup{3.8}
& 81.62\greenup{3.99} \\

AgentDropout
& 90.0\greenup{3.0}
& 51.6\greenup{4.0}
& 98.3\greenup{1.1}
& 91.0\greenup{2.5}
& 78.0\greenup{5.6}
& 75.6\greenup{2.5}
& 80.75\greenup{3.12} \\

\rowcolor{gray!10}
G-Designer
& 91.5\greenup{4.5}
& 52.4\greenup{4.8}
& 98.3\greenup{1.1}
& 91.5\greenup{3.0}
& 79.6\greenup{7.2}
& 76.9\greenup{3.8}
& 81.70\greenup{4.07} \\

ARG-Designer
& 92.5\greenup{5.5}
& 54.0\greenup{6.4}
& \underline{98.9}\greenup{1.7}
& 92.5\greenup{4.0}
& 80.0\greenup{7.6}
& \underline{77.5}\greenup{4.4}
& 82.57\greenup{4.94} \\

\rowcolor{gray!10}
TopoDIM
& 93.0\greenup{6.0}
& 55.0\greenup{7.4}
& \underline{98.9}\greenup{1.7}
& 92.5\greenup{4.0}
& \underline{81.0}\greenup{8.6}
& \underline{77.5}\greenup{4.4}
& 82.98\greenup{5.35} \\

GTD
& \underline{93.5}\greenup{6.5}
& \underline{55.5}\greenup{7.9}
& 98.2\greenup{1.0}
& \underline{93.0}\greenup{4.5}
& 80.4\greenup{8.0}
& \underline{77.5}\greenup{4.4}
& \underline{83.02}\greenup{5.39} \\

\midrule

\rowcolor{lightpurple}
\textbf{Codebook Agent (ours)}
& \textbf{94.8}\greenup{7.8}
& \textbf{56.5}\greenup{8.9}
& \textbf{99.4}\greenup{2.2}
& \textbf{95.4}\greenup{6.9}
& \textbf{83.5}\greenup{11.1}
& \textbf{78.1}\greenup{5.0}
& \textbf{84.62}\greenup{6.99} \\

\bottomrule
\end{tabular}
}

\caption{
\textbf{Main accuracy results (\%)} on reasoning and code-generation benchmarks.
HumanEval and MBPP report Pass@1.
Methods are grouped into single-agent prompting, multi-agent collaboration, and learned agent or topology design.
Deltas are absolute accuracy points relative to Vanilla; $\blacktriangle$\,/\,$\blacktriangledown$ mark improvements and decreases.
\textbf{Bold} is best in each column; \underline{underline} is second best.
All runs use gpt-4o-mini under the protocol of Section~\ref{sec:setup}.
}
\label{tab:main}
\end{table*}

\section{Experiments}
We design the evaluation to stress-test the three claims of the introduction with four questions:
(Q1)~Does amortized \method{} match or beat iterative query-conditioned designers on accuracy, or does indexing a short codebook trade quality for speed?
(Q2)~Is the useful design space truly a short list---do fixed topologies already sit within a point of generated ones, and does extra codebook capacity go unused?
(Q3)~Does an execution-grounded MLP over $\mathrm{vec}(A)$ cut tokens where the incumbent edge-count GNN cannot rank homogeneous teams, and is random selection enough?
(Q4)~Does one-pass codebook generation deliver the latency and token savings of Figure~\ref{fig:intro} without sacrificing transfer across LLM backends?

\subsection{Setup}
\label{sec:setup}
\textbf{Benchmarks.} Reasoning: GSM8K~\citep{cobbe2021gsm8k}, MATH~\citep{hendrycks2021math}, MultiArith~\citep{roy2015multiarith}, SVAMP~\citep{patel2021svamp}. Code: MBPP~\citep{austin2021mbpp}, HumanEval~\citep{chen2021humaneval} (Pass@1). Transfer: MMLU~\citep{hendrycks2021mmlu} under Qwen-3-8B (Table~\ref{tab:new-results}). Evaluation uses the fixed test splits of GSM8K (1314), MATH (500), MultiArith (180), SVAMP (1000), MBPP (500), HumanEval (160), and MMLU (1530); design-axis accuracy/token numbers use 200 tasks per benchmark (full MultiArith/HumanEval where smaller).
\textbf{Teams.} Math-format: $4{\times}$ MathSolver with FinalRefer. Code: $4{\times}$ Programming Expert with FinalRefer. MMLU: $3{\times}$ Knowledgeable Expert. Homogeneous profiles are the default (six of seven design-axis settings); HumanEval also runs a heterogeneous four-role team. Backbone gpt-4o-mini (temp.\ $0.7$, top-$p$ $1.0$, max tokens $1024$) unless stated; embeddings are frozen all-MiniLM-L6-v2 ($d{=}384$).

\textbf{Baselines.} Single-agent: Vanilla, CoT~\citep{wei2022cot}, SC-CoT~\citep{wang2023selfconsistency}. Multi-agent: LLM-Debate~\citep{du2024debate}, LLM-Blender~\citep{jiang2023llmblender}, DyLAN~\citep{liu2024dylan}. Learned design: GPTSwarm~\citep{zhuge2024gptswarm}, ADAS~\citep{hu2025adas}, AFLOW~\citep{zhang2025aflow}, MaAS~\citep{zhang2025maas}, AgentDropout~\citep{wang2025agentdropout}, G-Designer~\citep{zhang2025gdesigner}, ARG-Designer~\citep{li2026argdesigner}, TopoDIM~\citep{sun2026topodim}, GTD~\citep{jiang2026gtd}.
\textbf{Metrics.} Accuracy (Pass@1 for code); median topology-generation latency (3 warmup ${+}$ 50/100 timed calls on one GPU); mean LLM tokens/query; end-to-end wall clock. Single evaluation run per configuration.

\subsection{Main Results}
\label{sec:main-results}
\begin{figure}[t]
\centering
\includegraphics[width=\columnwidth]{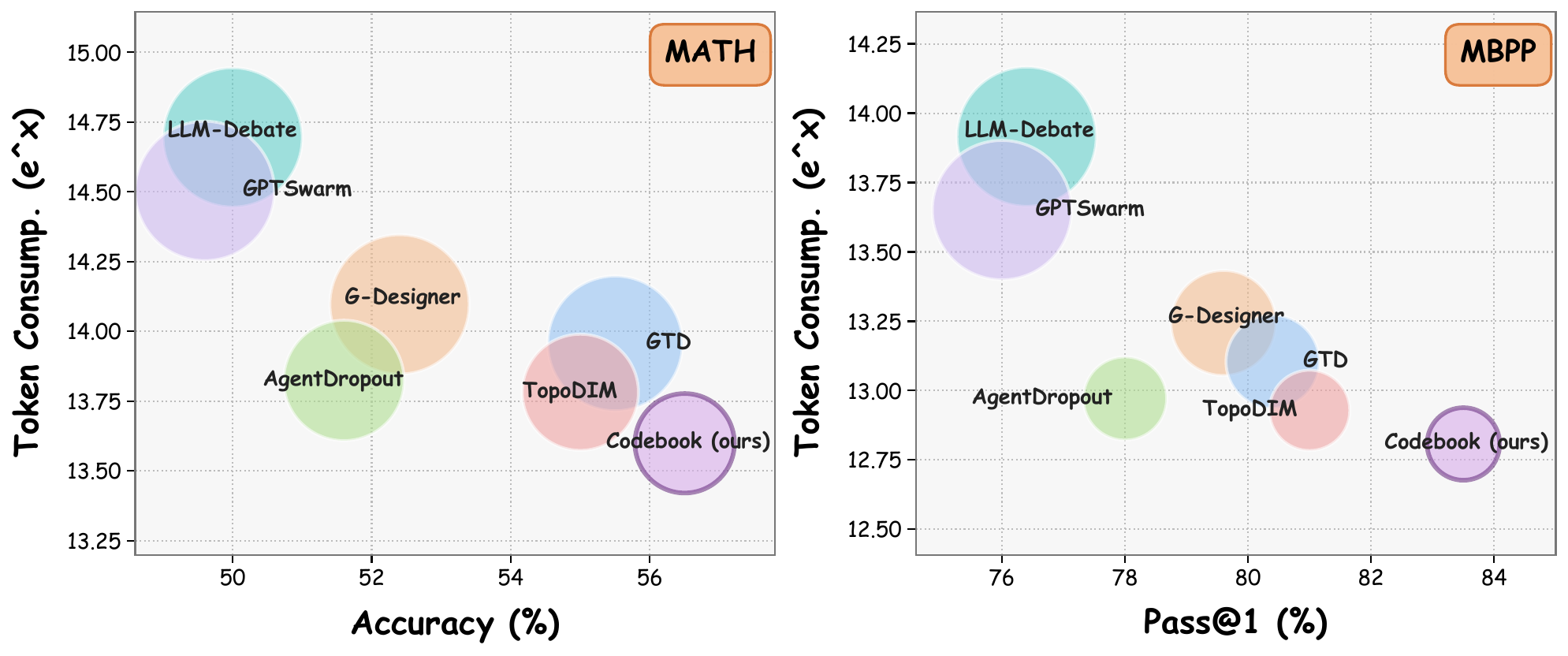}
\caption{\textbf{Accuracy--token trade-off on MATH and MBPP.}
Each bubble is one method: $x$ is accuracy, $y$ is $\ln(\mathrm{tokens}\times n)$ so that Token Consump.\ $=e^{y}$, and area scales with mean tokens per query. Lower-right is better; Codebook Agent sits nearest that corner on both benchmarks.}
\label{fig:tokens}
\end{figure}

Table~\ref{tab:main} reports gpt-4o-mini accuracy; Table~\ref{tab:new-results} the Qwen-3-8B transfer; Figures~\ref{fig:efficiency} and~\ref{fig:tokens} the latency and token frontiers. We highlight four observations corresponding to Q1--Q4.

\smallskip
\noindent\textbf{(Q1) Amortized selection clears the prior designer frontier; it does not trade accuracy for a codebook.}
Table~\ref{tab:main} puts \method{} best on every column at 84.62 average: $+6.99$ over Vanilla (77.63), $+4.44$ over DyLAN (80.18), and $+1.60$ over the strongest prior topology designer GTD (83.02). The Vanilla gap is smallest on MultiArith ($+2.2$; near ceiling) and largest on MBPP ($+11.1$). Against GTD the gains are spread, not concentrated: $+1.3$ GSM8K, $+1.0$ MATH, $+1.2$ MultiArith, $+2.4$ SVAMP, $+3.1$ MBPP, $+0.6$ HumanEval. Reading by family: single-agent prompting floors the high 70s; multi-agent collaboration lifts into the low 80s under hand-designed structure; query-conditioned generators (G-Designer through GTD) form the prior frontier at ${\sim}83$. \method{} sits above that frontier on all six benchmarks---so indexing a short list of graphs is not a speed concession; it matches or exceeds the iterative adjacency search it replaces.

\smallskip
\noindent\textbf{(Q2) The useful design space is a short list, and which fixed graph wins is not knowable a priori.}
Fixed topologies need no designer, so they measure how much of the payoff lives in the choice itself. Across seven settings the best fixed family (fully connected / chain / star) stays within 1.4 accuracy points of every generated topology we measure, at comparable tokens; on MATH, fully connected is above every generated configuration. Capacity spent modeling the adjacency manifold is therefore spent where the problem does not live---the premise of Axis~A as a codebook. The choice is nonetheless load-bearing: which fixed graph wins changes with the setting (fully connected on MATH, star on homogeneous HumanEval, chain on MMLU), with gaps up to 4.3 accuracy points and a $2.9{\times}$ token factor (chain vs.\ fully connected on MATH). Identifying the winner for a new setting requires real LLM executions---the same cost class as our one-time 300-record collection---so a designer earns its keep by amortizing that selection per query. Figure~\ref{fig:ablation}(b) closes the loop on capacity: as $K$ grows from 4 to 64, accuracy moves by at most 1.5 points (GSM8K) / 2.5 (HumanEval), and for every $K{\ge}8$ the encoder uses at most six codes. Collapse is a property of reward-surviving topologies, not of an undersized quantizer.

\smallskip
\noindent\textbf{(Q3) Measured-token MLP scoring is where cost is won; the incumbent GNN cannot rank anonymous teams.}
We hold Axis~A (codebook) and the candidate set fixed, and swap only Axis~B. On six homogeneous settings Eq.~\eqref{eq:incumbent} is constant in $A$, so every candidate in the short list receives the same score; the edge-count head is then the only separator, and because $|E|$ correlates with tokens at $r\approx -0.4$, it points toward expensive graphs. Empirically this incumbent \emph{scorer}, reranking our codebook candidates, costs 1711 tokens/query on GSM8K and 918 on homogeneous HumanEval---worse than uniform random over the same candidates (1249 / 750), and not comparable to the full incumbent \emph{pipeline} of Q4. Heterogeneous HumanEval is the exception on accuracy (incumbent 78.8 vs.\ ours 78.1); our largest token saving ($33.2\%$) is also there. Replacing our proxy with random, or dropping rerank and taking the predictor's top-1 (Figure~\ref{fig:ablation}c), shows the ranking is load-bearing for cost: random costs $23$--$35\%$ more tokens at equal or lower accuracy; top-1 is already cheap (913 tokens on GSM8K) and rerank adds $0.6$--$1.0$ accuracy points. The chain fixed topology is the extreme positive control for the inverted surrogate: sparsest connected, most token-expensive on every math-format benchmark.

\smallskip
\noindent\textbf{(Q4) One-pass generation is ${>}120{\times}$ faster and $22$--$33\%$ cheaper in tokens, and the ordering transfers.}
Figure~\ref{fig:efficiency} places MATH accuracy against median generation latency: iterative generators sit at 301--396\,ms; \method{} at 2.4--2.5\,ms ($125$--$158{\times}$). The gap is structural---$TK{=}250$ sparse GNN evaluations vs.\ one predictor pass, ${\le}M$ decodes, and one batched MLP call (Eq.~\eqref{eq:infer})---and design falls below $0.1\%$ of end-to-end latency. Figure~\ref{fig:tokens} shows the accuracy--token frontier on MATH and MBPP; \method{} occupies the lower-right corner. Against the full incumbent pipeline (iterative decoder ${+}$ edge-count GNN), mean tokens drop $21.9$--$33.2\%$ (e.g.\ $1239{\to}927$ GSM8K, $2304{\to}1611$ MATH, $699{\to}546$ HumanEval-homo, $624{\to}417$ HumanEval-het), and wall clock follows (MATH $43.7{\to}27.4$\,min; MMLU $12.7{\to}7.9$\,min). Table~\ref{tab:new-results} repeats the accuracy comparison with Qwen-3-8B: absolute numbers fall, but \method{} still leads at 74.0 vs.\ GTD 72.7 / G-Designer 72.1, with MATH $+1.7$ and MMLU $+1.9$ over GTD---so the codebook is not an artifact of one API.

\smallskip
\noindent\textbf{Overall.}
Q1--Q4 close the loop opened in the introduction: when reward-surviving topologies are few, accuracy is preserved by selecting among them (Q1--Q2), cost is won by scoring measured tokens rather than $|E|$ on anonymous teams (Q3), and amortization makes that selection essentially free at test time while transferring across backends (Q4). The ablations below isolate each mechanism under controlled Axis~A/B swaps.

\subsection{Ablations}
\label{sec:ablations}
\begin{figure*}[t]
\centering
\includegraphics[width=\textwidth]{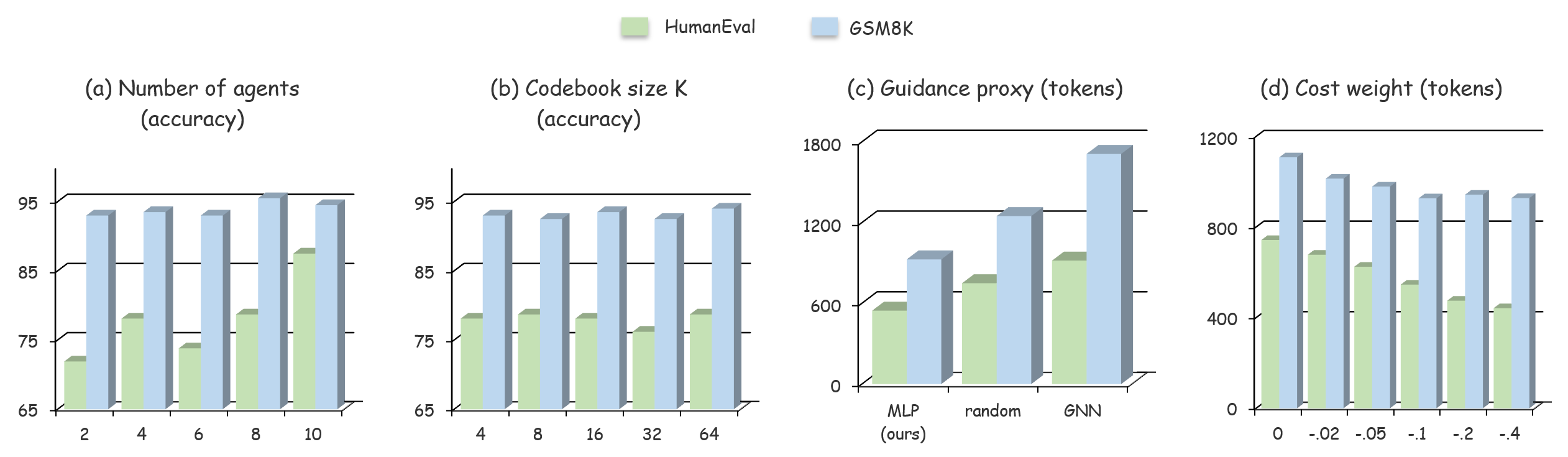}
\caption{\textbf{Ablations on GSM8K and homogeneous HumanEval.}
(a)~Accuracy versus team size $N$ (retrained per $N$ on the same training budget).
(b)~Accuracy versus codebook size $K$; used codes saturate near six for every $K{\ge}8$.
(c)~Mean tokens per query for MLP, random, and GNN reranking under a fixed candidate set.
(d)~Mean tokens versus the rerank cost weight $\lambda_{\mathrm{cost}}$, with default $-0.1$.}
\label{fig:ablation}
\end{figure*}

We ablate the mechanisms behind Eqs.~\eqref{eq:vq}--\eqref{eq:infer} and test whether the main-table ordering survives a backbone change. Figure~\ref{fig:ablation} varies four controls on GSM8K and homogeneous HumanEval under a fixed training budget; Table~\ref{tab:new-results} repeats the accuracy comparison with Qwen-3-8B agents.

\smallskip
\noindent\emph{(a)~Team size $N$.}
Figure~\ref{fig:ablation}(a): retrain per $N\in\{2,\dots,10\}$ on the same 50-task budget. GSM8K stays flat ($93.0$--$95.5$); HumanEval rises $71.9{\to}87.5$. Coding benefits from a wider expert pool, but the same index-and-select pipeline tracks that gain without architectural change---so the method is not a four-node trick, and topology selection remains useful as the team grows.

\smallskip
\noindent\emph{(b)~Codebook size $K$.}
Figure~\ref{fig:ablation}(b): $K\in\{4,\dots,64\}$ moves accuracy by ${\le}1.5$ (GSM8K) / $2.5$ (HumanEval). For every $K{\ge}8$ the encoder uses at most six codes---extra capacity stays idle under EMA with dead-code reset---so the design space, not the quantizer, is what collapsed. Searching the full $N{\times}N$ adjacency at test time cannot be justified by needing more capacity. The used-code count plateaus near the same six survivors across both benchmarks, which is why enlarging $K$ past 16 yields almost no accuracy return under a matched training budget.

\smallskip
\noindent\emph{(c)~Rerank signal.}
Figure~\ref{fig:ablation}(c): same Top-$M$ candidates; only the selector changes. Tokens on GSM8K / HumanEval are MLP $927$/$546$, random $1249$/$750$, GNN $1711$/$918$, with accuracy within $1.5$ points---the ranking buys cost, not large Pass@1. The GNN is worse than random because on homogeneous teams Eq.~\eqref{eq:incumbent} is constant in $A$ and its $|E|$ head favors expensive graphs ($r\approx -0.4$ with tokens). Only the MLP, reading $\mathrm{vec}(A)$ and regressing measured $\tilde{\tau}$, separates candidates on the true objective. Keeping the short list fixed isolates the critic: any token gap here is attributable to Eq.~\eqref{eq:infer} alone.

\smallskip
\noindent\emph{(d)~Cost weight $\lambda_{\mathrm{cost}}$.}
Figure~\ref{fig:ablation}(d): $\lambda{=}0$ is costliest ($1108$/$743$ tokens) with no accuracy gain. Tokens fall to $927$ (GSM8K) and $442$ (HumanEval at $-0.4$); accuracy peaks at $-0.1$/$-0.2$. Default $-0.1$ sits on the frontier: the proxy must both see structure and be asked to care about cost. The prior already prefers cheap codes (top-1: $913$ on GSM8K), and rerank adds $0.6$--$1.0$ accuracy points. Pushing $\lambda$ more negative continues to cut tokens but begins to trade accuracy, so $-0.1$ is a deliberate operating point rather than an unconstrained minimum-cost choice.

\smallskip
\noindent\emph{Qwen-3-8B transfer} (Table~\ref{tab:new-results}).
Same teams and 300-record protocol, backbone swapped to Qwen-3-8B. Absolute numbers drop (CoT avg.\ $67.5$), but \method{} still leads every column ($92.7$/$63.5$/$65.8$, avg.\ $74.0$) over GTD ($72.7$) and G-Designer ($72.1$). MATH ($+1.7$ vs.\ GTD) and MMLU ($+1.9$; three-expert team) show the gain is not confined to saturated arithmetic or four MathSolvers. Design never calls the agent LLM (Eq.~\eqref{eq:infer}), so swapping the backbone changes only the rewards in $\mathcal{D}$---the short-list recipe is not an API artifact. Relative ordering among adaptive designers is preserved under the weaker model, which suggests that the amortized index, not gpt-4o-mini idiosyncrasies, drives the main-table gaps.

\begin{table}[htbp]
\centering
\renewcommand\arraystretch{1.2}

\vspace{-0.5em}
\resizebox{\columnwidth}{!}{%
\begin{tabular}{@{}lcccc@{}}
\toprule
\rowcolor{gray!15} 
\textbf{Method} & \textbf{GSM8K} & \textbf{MATH} & \textbf{MMLU} & \textbf{Avg.} \\
\midrule
CoT
& 87.8
& 55.0
& 59.6
& 67.5 \\

GPTSwarm
& 90.9\greenup{3.1}
& 60.0\greenup{5.0}
& 63.5\greenup{3.9}
& 71.5\greenup{4.0} \\

\rowcolor{gray!10}
MaAS
& 91.2\greenup{3.4}
& 60.4\greenup{5.4}
& 63.9\greenup{4.3}
& 71.8\greenup{4.3} \\

G-Designer
& 91.5\greenup{3.7}
& 60.2\greenup{5.2}
& 64.6\greenup{5.0}
& 72.1\greenup{4.6} \\

\rowcolor{gray!10}
GTD
& 92.3\greenup{4.5}
& 61.8\greenup{6.8}
& 63.9\greenup{4.3}
& 72.7\greenup{5.2} \\

\midrule
\rowcolor{lightpurple}
\textbf{Ours}
& \textbf{92.7\greenup{4.9}}
& \textbf{63.5\greenup{8.5}}
& \textbf{65.8\greenup{6.2}}
& \textbf{74.0\greenup{6.5}} \\
\bottomrule
\end{tabular}%
}
\vspace{0.5em}
\caption{\textbf{Qwen-3-8B transfer.}
Accuracy (\%) on GSM8K, MATH, and MMLU under the same agent teams and protocol as Section~\ref{sec:setup}, with Qwen-3-8B replacing gpt-4o-mini.
Avg.\ is the three-benchmark mean; deltas are from CoT; bold marks each column's best.}
\vspace{-0.5em}
\label{tab:new-results}
\end{table}

\section{Conclusion}

We find that effective multi-agent topologies form a short list, that denser graphs need not cost fewer tokens, and that message-passing critics can miss structure on homogeneous teams. Accordingly, Codebook Agent indexes a small discrete set of topologies and selects among them with a lightweight proxy (Eqs.~\eqref{eq:vq}--\eqref{eq:infer}), leading all six benchmarks and the Qwen-3-8B transfer while cutting design latency to 2.4\,ms and LLM tokens by 21.9--33.2\%.

\clearpage
\bibliography{reference}

\end{document}